\documentclass[runningheads,a4paper]{llncs}

\usepackage{amssymb}
\usepackage{graphicx}
\usepackage{amsmath}
\usepackage{multirow}

\usepackage{url}
\newcommand{\keywords}[1]{\par\addvspace\baselineskip
\noindent\keywordname\enspace\ignorespaces#1}
\newcommand{\s}[1]{#1}

\begin{document}

\mainmatter

\title{A Comparison of Random Forests and Ferns on Recognition of Instruments in Jazz Recordings}

\titlerunning{A Comparison of Random Forests and Ferns on Recognition of Instruments}

\author{Alicja A. Wieczorkowska\inst{1}%
\and Miron B. Kursa\inst{2}}

\authorrunning{A.A. Wieczorkowska and M.B. Kursa}

\institute{Polish-Japanese Institute of Information Technology,\\
Koszykowa 86, 02-008 Warsaw, Poland\ \\ \email{alicja@poljap.edu.pl}\ \\ \and Interdisciplinary Centre for Mathematical and
Computational Modelling (ICM), University of Warsaw, \\ Pawi\'{n}skiego 5A,
02-106 Warsaw, Poland\  \\
\email{M.Kursa@icm.edu.pl}}

\toctitle{A Comparison of Random Forests and Ferns on Recognition of Instruments in Jazz Recordings}
\tocauthor{Alicja A. Wieczorkowska and Miron B. Kursa}
\maketitle

\pagenumbering{gobble}

\begin{abstract}
In this paper, we first
apply random ferns for classification of real music recordings of a jazz band. No initial segmentation of audio data is assumed, i.e., no onset, offset, nor pitch data are needed. The notion of random ferns is described in the paper, to familiarize the reader with this classification algorithm, which was introduced quite recently and applied so far in image recognition tasks. The performance of random ferns is compared with random forests for the same data. The results of experiments are presented in the paper, and conclusions are drawn.
\\

\end{abstract}

\begin{center}
\textit{This is a preprint of a paper accepted for the ISMIS 2012 conference.}
\end{center}

\section{Introduction}

The pleasure of listening to music can be very enjoyable, especially if our favorite instruments are playing in the piece of music we are listening to. Therefore, it is desirable to have a tool to find melodies played by a specified instrument. The task of automatic identification of an instrument, playing in a given audio segment, lies within the area of interest of Music Information Retrieval. This area has been broadly explored last years \cite{MIR}, \cite{shen}, and as a result we can enjoy finding pieces of music through query-by-humming \cite{midomi}, and identify music through query-by-example, including excerpts replayed on mobile devices \cite{shazam}, \cite{trackid}. However, recognition of instruments in real polyphonic recordings is still a challenging task (see e.g. \cite{her1}, \cite{kita}, \cite{ISMIS11}).

In this paper, we address the recognition of plural instruments in real music recordings of a jazz band, and
our goal is to identify possibly all instruments playing in each audio frame; polyphony in these recordings reaches 4 instruments. Identification of instruments is performed in short frames,
with no assumption on onset (start) nor offset (end) time, nor pitch etc., which is often the case in similar research, thus our methodology requires no preprocessing  nor initial segmentation of the data, and the computation can be fast.

Random ferns are classifiers introduced in 2007 \cite{Ozuysal2007} and named as such in 2008 \cite{Ozuysal2008}. This classification method combines features of decision trees and Bayesian classifiers. Random ferns have been applied so far in image classification tasks, including video data \cite{Bosch2007}, \cite{Oshin2009}, and they have also been adjusted to be used on low-end embedded platforms, such as mobile phones \cite{Wagner2010}. Since many audio applications are used in mobile environment, it is advisable to consider such platforms as well. This is why we decided to use random ferns. Additionally, we would like to compare the performance of Random Ferns (RFe) with Random Forests (RFo), which yielded quite good results in our previous research \cite{ISMIS11}, \cite{Praga2009}, \cite{rsctc}. RFe are simpler and more computationally efficient than RFo \cite{JSS}.
We want to use a simpler algorithm because, as more computationally efficient, it can possibly be applied to be used on mobile devices, with limited computational power (utilizing slower CPUs and working on battery power).
We hope that the accuracy of RFe is not much worse, and therefore it is worth using them and possibly implement on mobile devices, to get quick results without communication with a cloud for cloud computing (which is an option which can be chosen for low-end platforms), thus achieving low latency.
Also, such a method would be useful for massive calculations for indexing purposes, e.g. in archives, to achieve fast computation and get quick results which are a good approximation of the results that would be obtained using more computationally expensive search.

\section{Classifiers}

The classifiers applied in our research include random ferns and random forests. RFo performed quite well in the research on instrument identification we performed before \cite{ISMIS11}, but their training is time consuming, whereas the training of RFe is faster. The computational complexity of classification performed using the pre-trained classifiers is similar (linearly proportional to the number of trees/ferns and to their average height), but in the case of ferns there is less branching and memory accesses which should yield faster classification.

\subsection{Random Forests}

RFo is a classifier consisting of a set of weak, weakly correlated and non-biased decision trees, constructed using a
procedure minimizing bias and correlations between individual
trees \cite{Breiman}. Each tree is built using a different $N$-element bootstrap
sample of the
training $N$-element set. The elements of the bootstrap sample are drawn
with replacement from the original set, so
 roughly 1/3 (called \textit{out-of-bag}) %
 of the training data are not used in the bootstrap
sample for any given tree.
For a $P$-element feature vector, $K$ attributes (features) are randomly selected
at each stage of tree building, i.e. for
each node of any particular tree in RFo ($K<P$, often
$K=\sqrt{P}$).
Gini impurity criterion (GIC) is applied to find the best split on these $K$ attributes. GIC measures how often an element would be incorrectly labeled if randomly labeled according to the distribution of labels in the subset; the best split minimizes GIC.

Each tree is grown to the largest
extent possible, without pruning.
By repeating this randomized
procedure $N_{t}$ times, a collection of $N_{t}$ trees is obtained, constituting a RFo.
Classification of an object is done by simple voting of all trees.
In this work, the RFo implementation from the R \cite{R} package randomForest \cite{Liaw2002} was used.

The computational complexity $Ct_{Fo}$ of training a RFo is
\begin{equation}
Ct_{Fo} = N_t\cdot N_o\cdot \log N_o\cdot K \;  , 
 \end{equation}
where $N_{o}$ is the number of objects, $K$ is the number of attributes tested for each split and $N_{t}$ is the number of trees in the forest;
the computational complexity
\begin{equation}
Cc_{Fo} = N_{t}\cdot h_{t} \;
\end{equation}
where $h_{t}$ is the average height of a tree in the forest.

\subsection{Random Ferns}

\begin{figure}
   \centering
   \includegraphics[width=0.7\textwidth]{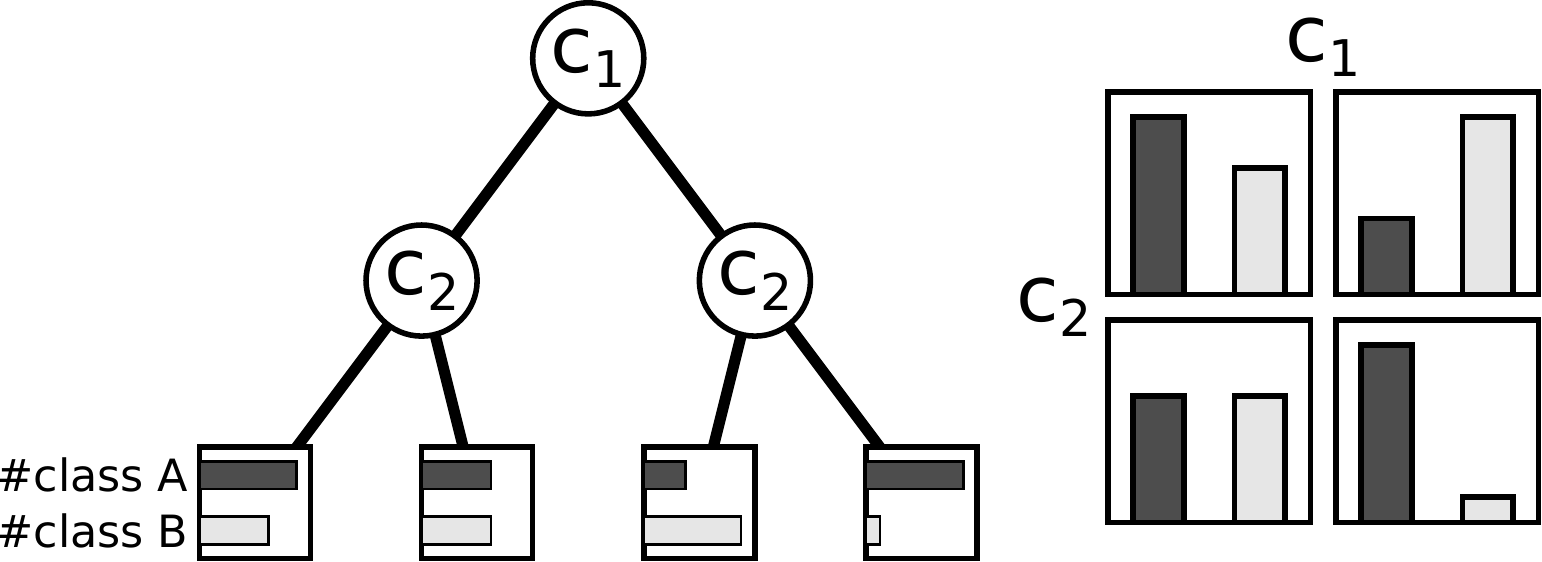}
   \caption{An example of a fern of depth 2 trained on a binary classification problem (\textit{left}).
   Splits on each level are based on the same criterion ($C_i$), thus the fern tree is equivalent to a 2-dim array (\textit{right}).
   The leaf nodes contain the counts of objects of each class instead of just the names of dominating classes, as in classic decision trees.
   }
   \label{fig:fern}
\end{figure}

A fern is defined as a simplified binary decision tree of a fixed height $D$ (called a \textit{depth} of a fern) and with a requirement that all splitting criteria at a certain depth $i$ ($C_i$) are the same.
Each leaf node of a fern stores the distribution of classes over objects that are directed to this node.
This way a fern can be perceived as a $D$-dimensional
array of distributions, indexed by a vector of $D$ splitting criteria values, see Figure~\ref{fig:fern}.

\begin{figure}
   \centering
   \includegraphics[width=0.55\textwidth]{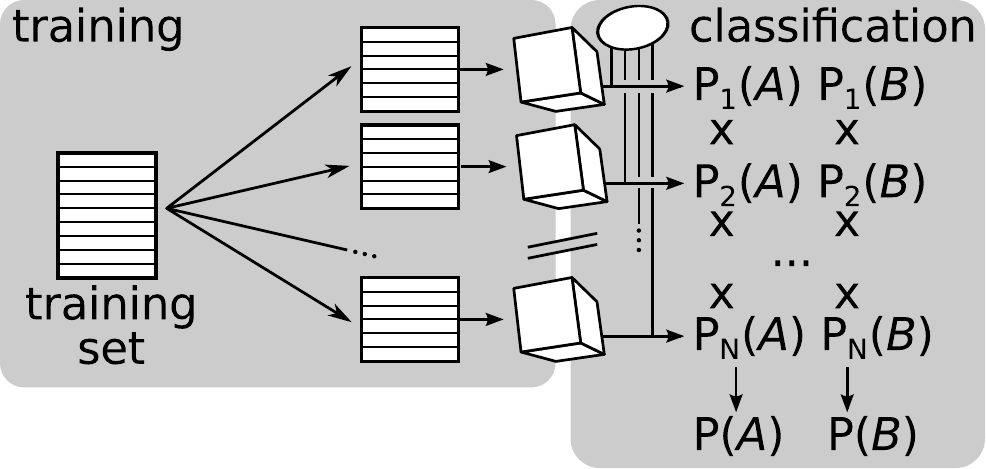}
   \caption{Training and
   classification using
   a fern forest
   for a binary classification problem.
     \textit{Bags} are drawn from the training data,
   and
   used for building individual ferns, represented here as cubes (\textit{left}).
   When a new object (represented as an ellipse) is
   classified, each fern in the
   forest returns a vector of class probabilities; they are combined by a simple multiplication and the class
   scoring maximal probability is returned (\textit{right}).
     }
   \label{fig:ferns}
\end{figure}

The fern forest is a collection of $N_{f}$ ferns.
When
classifying a new object, each fern in a forest returns a vector of probabilities that this object belongs to particular decision classes.
Ferns are treated as independent, thus all those vectors are combined by simple multiplication and the final
classification results for the forest is a class which gets the highest probability, see Figure~\ref{fig:ferns}.

While the original RFe implementation \cite{Ozuysal2007,Ozuysal2008} was written for a
problem of object detection in images, we use the RFe generalization implemented in the R \cite{R} package rFerns \cite{JSS}; it trains the fern forest model in the following way.

First, $N$ intermediate training sets called \textit{bags} are created
by drawing objects with replacement from the training set, each bag being of the same size as the original set.
Next, each bag is used to train a fern. All $D$ splits are created purely at random; an attribute is randomly selected and then the splitting threshold is set as a mean of two randomly selected values of this attribute\footnote{In this work we have used only numerical descriptors of sound, thus the description of treating ordinal and categorical attributes is omitted.}.
The distributions of classes in leafs are calculated on a bag  with adding 1 for each class (i.e. with a Dirichlet prior); this way the problem of undefined distributions in leafs containing no objects is resolved.

The computational complexity $Ct_{Fe}$ of training a Rfe model is
\begin{equation}
Ct_{Fe} = 2^D\cdot N_f \cdot N_o \;  , 
\end{equation}
where $D$ --- depth of ferns, $N_o$ --- number of objects, $N_f$ --- number of ferns;
the computational complexity $Cc_{Fe}$ of classifying one sample is
\begin{equation}
Cc_{Fe} = D\cdot N_f \;  .
\end{equation}

\section{Sound Parameterization}

The identification of musical instruments is performed for short frames of audio data, which are parametrized before applying classifiers for training or testing. No assumptions on audio data segmentation or pitch extraction have been made.
Therefore, no multi-pitch extraction is needed, thus avoiding possible errors regarding labeling particular
sounds in polyphonic recording with the appropriate pitches.
The feature vector consists of basic features, describing properties of an audio frame of 40 ms, and additionally difference features, calculated as the difference between the given feature but calculated for a 30 ms sub-frame starting from the beginning of the frame and a 30 ms sub-frame starting 10 ms later.
Identification of instruments is performed frame by frame, for consequent frames, with 10 ms hop size. Fourier transform was used to calculate spectral features, with Hamming
window. Most of the features we applied represent
MPEG-7 low-level audio descriptors, which are often used in audio research \cite{MPEG-7}. Our feature vector consists of the following 91 parameters \cite{ISMIS11}:

\begin{itemize}
\item \textit{Audio Spectrum Flatness}, \textit{flat}$_{1}, \ldots,\; $\textit{flat}$_{25}$ --- a multidimensional
parameter describing the flatness property of the power spectrum
within a frequency bin for selected bins; 25 out of 32 frequency
bands were used; 
\item \textit{Audio Spectrum Centroid} --- the power weighted average of the
frequency bins in the power spectrum; coefficients are scaled to an octave scale anchored at 1 kHz \cite{MPEG-7};
\item \textit{Audio Spectrum Spread} --- 
RMS (root mean square) value of the deviation of
the log frequency power spectrum 
wrt. \textit{Audio Spectrum Centroid}
\cite{MPEG-7};
\item \textit{Energy} --- energy (in log scale) of the spectrum of the parametrized sound;
\item \textit{MFCC} --- a vector of 13 mel frequency cepstral coefficients. The cepstrum was calculated as the logarithm of
the magnitude of the spectral coefficients, and then transformed to the mel scale, to better reflect properties of the human perception of frequency. 
24 mel filters were applied, and the obtained results were transformed 
to 
12 coefficients. The 
$13^{th}$ coefficient is the 0-order coefficient of MFCC, corresponding to the logarithm of the energy
\cite{MFCC};
\item \textit{Zero Crossing Rate}; a zero-crossing is a point where the sign of the time-domain representation of the sound
wave changes; 
\item \textit{Roll Off} --- the frequency below which an experimentally
chosen percentage equal to 85\% of the accumulated magnitudes of
the spectrum is concentrated; parameter originating from
speech recognition, where it is applied to distinguish between voiced and unvoiced speech;
\item \textit{NonMPEG7 - Audio Spectrum Centroid} --- a linear scale version of \textit{Audio Spectrum Centroid};
\item \textit{NonMPEG7 - Audio Spectrum Spread} --- a linear scale version of \textit{Audio Spectrum Spread};
\item changes (measured as differences) of the above features
    for a 30 ms sub-frame of the given 40 ms frame (starting from the beginning of this frame) and the next 30 ms sub-frame (starting with 10 ms shift), calculated for all the features shown above;
\item \textit{Flux} --- the sum of squared
differences between the magnitudes of the DFT points calculated for the starting and ending 30 ms sub-frames within the main 40 ms frame; this feature by definition describes changes of magnitude spectrum, thus it is not calculated in a static version.
\end{itemize}

Mixes of the left and right channel were taken if the audio signal was stereophonic. Since the recognition of instruments is performed on frame-by-frame basis, no parameters describing the entire sound are present in our feature vector. This feature set was already used for instrument identification purposes using RFo, requiring no feature selection \cite{ISMIS11},
and yielded good results, so we decided to use this feature set
in both RFo and RFe classification.

\subsection{Audio Data}\label{Audio}

The audio data we used for both training and testing represent recordings in 44.1kHz/16-bit format.
Training was based on three repositories of single, isolated sounds of musical instruments, namely McGill University Master Samples \cite{Opo_Wap}, The
University of Iowa Musical Instrument Samples \cite{IOWA}, and RWC Musical Instrument Sound Database
\cite{goto}. Clarinet, trombone, and trumpet sounds were taken from these repositories. Additionally, we used sousaphone sounds, recorded by R. Rudnicki in one of his recording sessions \cite{RR}, since no sousaphone sounds were available in the above mentioned repositories. Training data were in mono format in the case of RWC data and sousaphone, and stereo for the rest of the data.
The testing data originate from jazz band stereo recordings by R. Rudnicki
\cite{RR}, and include the following pieces played by clarinet, trombone, trumpet, and sousaphone (i.e., our target instruments):
\begin{itemize}
\item Mandeville by Paul Motian,
\item Washington Post March by John Philip Sousa, arranged by Matthew Postle,
\item Stars and Stripes Forever by John Philip Sousa, semi-arranged by Matthew Postle --- Movement no. 2 and Movement no. 3.
\end{itemize}
To prepare our classifiers to work on larger instrument sets, training data also included sounds of 5 other instruments that
can be encountered in jazz recordings: double bass, piano, tuba, saxophone, and harmonica. These sounds were added as additional sounds in training mixes with the target instruments.

\section{Methodology of Training of the Classifiers}\label{train}
The goal of training of our classifiers is to identify plural classes, each representing one instrument. We use a set of binary classifiers (RFe or RFo), where each set (which we call a \textit{battery}) is trained to identify whether a target instrument is playing in an audio frame or not. The target classes are clarinet, trombone, trumpet, and sousaphone, i.e. instruments playing in the analyzed jazz band recordings. The classifiers are trained to identify target instruments when they are accompanied by other instruments, and this is why we use mixes of instrument sounds as input data in training.

When preparing training data, we start with single isolated sounds of each target instrument. After removing starting and ending silence \cite{ISMIS11},  each file  representing the whole single sound is normalized so that the RMS value equals one.
Then we perform parameterization, and train a classifier to identify each instrument --- even when accompanied by other sound. Therefore, we perform training on 40 ms frames of instrument sound mixes, mixing from 1 to 4 randomly chosen instruments with random weights and then we normalize it again to get the RMS value equal to one.

The battery of one-instrument sensitive RFo or RFe classifiers is then trained. 3,000 mixes containing any sound of a given instrument are fed as positive examples, and 3,000 mixes containing no sound of this instrument are fed as negative examples. For N instruments we need N binary classifiers (N=4), each one trained to identify 1 instrument.
For RFe models, we have been training 1000 ferns of a depth of 10; for RFo, there were 1000 trees and $K$ was set to the default floor of square root of the number of attributes, namely 9.

\section{Experiments and Results}
The RFo and RFe classifiers, according to the procedure delineated in Section \ref{train}, were next used to identify instruments playing in jazz recordings, described in Section \ref{Audio}.
Ground-truth data were prepared through careful manual labelling \cite{ISMIS11}, based on initial recordings of each instrument track separately.

The accuracy was assessed via precision and recall scores.
These measures were weighted by the RMS of a given frame (differently than in our previous work \cite{ISMIS11}, where RMS was calculated for frames taken from instrument channels), in order to
 diminish the impact of softer frames, which are very hard to perform reasonable identification of instruments, because their loudness is near the noise level.
For this reason, our true positive score $T_{p}$ for an instrument $i$ is a sum of RMS of frames which are both annotated and 
classified as $i$. Precision is calculated by dividing $T_{p}$ by the sum of RMS of frames which are 
classified as $i$; respectively, recall is calculated by dividing $T_{p}$ by the sum of RMS of frames which are annotated as $i$.
As a general accuracy measure we have used F-score, defined as a harmonic mean of such precision and recall.
\begin{table}[htb]
\caption{Precision, recall and F-score of the classifiers for jazz band recordings. Each $M\pm S$ data entry represents mean $M$ and standard deviation $S$ over 10 replications of training and testing, accumulated over all target band instruments.}
\centering
\begin{tabular}{|c||c|c|c|}
\hline
Algorithm & Precision [\%]  & Recall [\%] & F-score [\%]  \\
\hline
\hline
\multicolumn{4}{|c|}{Mandeville}\\
\hline
RFe &    88.4$\pm$0.6  & \s{67$\pm$1} & 76.4$\pm$0.6\\
RFo & \s{92.7$\pm$0.2} &    63$\pm$1  & 75.2$\pm$0.7\\
\hline
\multicolumn{4}{|c|}{Washington Post}\\
\hline
RFe &     82.36$\pm$0.2  & 73$\pm$2 & 77$\pm$1\\
RFo &  \s{87.76$\pm$0.3} & 69$\pm$1 & 77.3$\pm$0.5\\
\hline
\multicolumn{4}{|c|}{Stars \& Stripes 2}\\
\hline
RFe &    79.8$\pm$0.4 & \s{72$\pm$1} & 76$\pm$1\\
RFo & \s{91$\pm$2}    &     68$\pm$1 & 78$\pm$1\\
\hline
\multicolumn{4}{|c|}{Stars \& Stripes 3}\\
\hline
RFe & 94.5$\pm$0.2 & \s{77$\pm$1} & 84.8$\pm$0.7\\
RFo & 94.4$\pm$0.3 &    74$\pm$1  & 83.1$\pm$0.9\\
\hline
\end{tabular}
\label{tab:predAccWhole}
\end{table}

While in this initial phase of the research we have used PC implementations of the classification algorithms, the timings have been performed on a single core of a Xeon E5620 Linux workstation. R version 2.15.0, rFerns version 0.3 and randomForest version 4.6-6 were used.

Both RFo and RFe are stochastic algorithms, so is the process of creating training sets for the battery.
Thus, to assess the stability of the results and make a fair comparison of methods, the whole procedure of creating training sets, training RFe and RFo batteries and testing them on a real recordings has been repeated 10 times.

\subsection{Comparison of Random Forests and Random Ferns}

The results of performance analysis of RFe and RFo models are given in Table~\ref{tab:predAccWhole}.
As one can see, for three pieces RFo had superior precision over that of RFe; on the other hand, ferns tend to provide better recall.
However, the overall performance of both classifiers measured with the F-score is similar for all pieces.

The detailed comparison of performance analysis of RFe and RFo models for particular instruments is given in Table~\ref{tab:predAccInstr}.
Sousaphone and trumpet are always quite precisely identified, whereas trombone usually yields lower precision in all pieces, and clarinet in one piece.
Recall is lower than precision, but still much improved comparing to our previous results \cite{ISMIS11}.
Again, quite high recall is obtained for sousaphone and is rather good for trumpet, whereas the worst recall is scored by RFo for trombone samples.

\begin{table}[htb]
\caption{Precision and recall of both methods on real music; data shown for each instrument independently.
The symbol $M\pm S$ denotes that given number has mean $M$ and standard deviation $S$ over 10 replications of training.}
\centering
\begin{tabular}{|c||c|c|c|c||c|c|c|c|}

\hline
& \multicolumn{4}{|c||}{Precision [\%]} & \multicolumn{4}{|c|}{Recall [\%]} \\
\hline
&clarinet  & sousaphone  & trombone  & trumpet   & clarinet  & sousaphone  & trombone & trumpet  \\
\hline
\hline
\multicolumn{9}{|c|}{Mandeville}\\
\hline
RFe & 91.5$\pm$0.2 & 98.3$\pm$0.2 & 76$\pm$2     & 89.0$\pm$0.2 &               70$\pm$2 & 67$\pm$1 & 71$\pm$2 & 59$\pm$2 \\
RFo & 91.4$\pm$0.2 & 98.6$\pm$0.3 & 87.3$\pm$0.6 & 90.8$\pm$0.2 &               65$\pm$4 & 80$\pm$2 & 46$\pm$2 & 58$\pm$2 \\
\hline
\multicolumn{9}{|c|}{Washington Post}\\
\hline
RFe & 80.9$\pm$0.4 & 92.2$\pm$0.7 & 63.6$\pm$0.4 & 92.5$\pm$0.5 &                79$\pm$3 & 76$\pm$3 & 61$\pm$2 & 73$\pm$2 \\
RFo & 85$\pm$1     & 93.2$\pm$0.7 & 70.3$\pm$0.8 & 96.4$\pm$0.6 &                67$\pm$4 & 88$\pm$1 & 46$\pm$2 & 72$\pm$3 \\
\hline
\multicolumn{9}{|c|}{Stars \& Stripes 2}\\
\hline
RFe & 48.4$\pm$0.4 & 99.4$\pm$0.1 & 78$\pm$2 & 97.8$\pm$0.3 &               81$\pm$2 & 70$\pm$4 & 58$\pm$2 & 77$\pm$2 \\
RFo & 62$\pm$6     & 99.4$\pm$0.1 & 91$\pm$2 & 99.9$\pm$0.1 &               53$\pm$5 & 94$\pm$1 & 31$\pm$3 & 67$\pm$3 \\
\hline
\multicolumn{9}{|c|}{Stars \& Stripes 3}\\
\hline
RFe & 96.9$\pm$0.6 & 99.2$\pm$0.2 & 88.6$\pm$0.6 & 94.7$\pm$0.2 &               92$\pm$2 & 62$\pm$4 & 61$\pm$2 & 88$\pm$1 \\
RFo & 95.4$\pm$0.4 & 99.7$\pm$0.1 & 87.8$\pm$0.7 & 94.7$\pm$0.1 &               80$\pm$5 & 83$\pm$4 & 50$\pm$2 & 88$\pm$1 \\
\hline
\end{tabular}
\label{tab:predAccInstr}
\end{table}

\begin{table}[htb]

\caption{Time taken by each battery of classifiers to annotate the whole testing piece.}

\begin{tabular}{|c||c|c|c|}

\hline
\multirow{2}{*}{Piece}& \multicolumn{2}{|c|}{Classification time [s]} & \multirow{2}{*}{Piece length [s]}\\

 & Random Ferns & Random Forest & \\
\hline \hline
Mandeville & 5.7 & 17.6 & 139.95 \\
Washington Post & 6.0 & 19.0 & 148.45\\
Stars \& Stripes 2 & 2.8 & 8.3 & 68.95\\
Stars \& Stripes 3 & 1.0 & 3.6 & 26.2\\
\hline
\end{tabular}
\centering
\label{tab:predTimes}
\end{table}

On average, RFe and RFo perform classification respectively 25x and 8x faster than the actual music speed; this
means RFe offer over 3x speed-up in comparison to RFo, see Table~\ref{tab:predTimes}.

\section{Summary and Conclusions}

Experiments presented in this paper show that identification of all instruments playing in real music recordings is possible using both RFo- and RFe-based classifiers, yielding quite good results.
We observed improved recall comparing to our previous research \cite{ISMIS11}; we improved here the RMS weighting, which was previously calculated for separate instrument channels, and in this work, the RMS of all channels together was used for weighting.
Our results still are worth improving, but the obtained recall (and precision) are satisfactory, because the task of identification of all instruments playing in a short segment is difficult, and is challenging also for human listeners.

The measured classification speed of RFe suggests that it is a promising method for performing real time annotation, even on low performance devices.

\subsubsection*{Acknowledgments.} This work has been partially financed by the National Science Centre, grant 2011/01/N/ST6/07035. Computations were performed at ICM, grant G48-6.
This project was also partially supported by the Research Center of PJIIT, supported by the Polish Ministry of Science and Higher Education.
The authors would also like to thank Dr. El\.{z}bieta Kubera from the University of Life Sciences in
Lublin for preparing the ground-truth data for initial experiments, and Rados\l{}aw Rudnicki from the University of York for preparing the jazz band recordings.

\end{document}